%% file: index.tex
\documentclass[sigconf]{acmart}

\usepackage{booktabs} 
\usepackage{graphicx,dblfloatfix}
\usepackage{subfig}

\copyrightyear{2019} 
\acmYear{2019} 
\setcopyright{acmlicensed}
\acmConference[GECCO '19 Companion]{Genetic and Evolutionary Computation
Conference Companion}{July 13--17, 2019}{Prague, Czech Republic}
\acmBooktitle{Genetic and Evolutionary Computation Conference Companion
(GECCO '19 Companion), July 13--17, 2019, Prague, Czech Republic}
\acmPrice{15.00}
\acmDOI{10.1145/3319619.3326878}
\acmISBN{978-1-4503-6748-6/19/07}

\begin{document}
\title[Benchmarking Surrogate-Assisted Genetic Recommender Systems]{Benchmarking\\Surrogate-Assisted Genetic Recommender Systems}

\author{Thomas Gabor and Philipp Altmann}
\affiliation{%
  \institution{LMU Munich}
}

\renewcommand{\shortauthors}{Gabor and Altmann}

\begin{abstract}
We propose a new approach for building recommender systems by adapting surrogate-assisted interactive genetic algorithms. 
A pool of user-evaluated items is used to construct an approximative model which serves as a surrogate fitness function in a genetic algorithm for optimizing new suggestions. The surrogate is used to recommend new items to the user, which are then evaluated according to the user's liking and subsequently removed from the search space.
By updating the surrogate model after new recommendations have been evaluated by the user, we enable the model itself to evolve towards the user's preferences. 

In order to precisely evaluate the performance of that approach, the human's subjective evaluation is replaced by common continuous objective benchmark functions for evolutionary algorithms. The system's performance is compared to a conventional genetic algorithm and random search. We show that given a very limited amount of allowed evaluations on the true objective, our approach outperforms these baseline methods.
\end{abstract}

%
%



\maketitle

\input{contents.tex}

\bibliographystyle{ACM-Reference-Format}
\bibliography{references} 

\end{document}

%% file: contents.tex
\section{Introduction}

During the last century, consumer behavior shifted from buying products to subscribing to services.
Platforms like Spotify or Netflix almost entirely replaced the need to buy CDs, DVDs or even digital copies.
However, besides offering an ultimate freedom of choice, this overwhelming variety also causes an information overload, leaving users with the problem of finding songs that match their taste.
Luckily, a vast amount of research has been put into developing and improving systems helping the user to deal with this overload by recommending items that are likely to match his or her taste.

In contrast to the most popular recommender method of collaborative filtering, our approach is not taking other users' opinions into account in order to increase the prediction precision.
Instead, we are using genetic algorithms to optimize suggestions with respect to a surrogate model, constructed from the currently evaluated items and updated after every suggestion.
As frequently updating the surrogate model is indirectly optimizing the model itself, the strategy we propose is able to adapt to changes in the user's taste and find the most viable suggestions, even though similar items might not have been evaluated yet.
Thus we rely on the ability of the surrogate model to suggest supposedly good items and then make drastic updates if the suggested items has not been as well-received.
We argue that this approach may be able to recommend more diverse items without the need to compare to or even access other users' data.
To show the viability of this concept from an algorithmic point of view, we simulate the user's taste through common benchmarking functions for evolutionary algorithms.
Motivated by the user-interactive scenario though, our experiments differ from common surrogate-assisted algorithms in goal of recommendation, i.e., we explicitly exclude any individuals that have already been evaluated by the true objective function from our optimization process, yielding a highly dynamic optimization process.

We first review some related work regarding recommender systems and surrogate-assisted genetic algorithms in Section~2.
In Section 3 we introduce the approach and provide more detailed explanations on the parameters and surrogate models used.
After that, we evaluate the concept by testing different settings and analyze the suitability of the different meta-models in Section 4. Finally, we sum up the findings, discuss limitations and prospects and show possibilities of future work in Section 5.

\section{Foundations and Related Work}

\subsection{Recommender Systems}
Research regarding recommender systems began in the mid-1990s \cite{Hill:1995df, Resnick:1994dq} with the motivation of providing useful suggestions to users in order to help them make choices in a space too overcrowded to be survey-able by a human.
By overcrowded spaces, we refer to item domains that consist of far more items than a user can compare or evaluate.
Also, the density of items in some areas of the given domain decreases their comparability and exacerbates the human selection.


In general, recommender systems rely on rating data provided by the users. In an effort to predict highly rated items, they use filtering methods to reduce the number of items that could be suggested and recommend new items to their users so that these are likely to match their tastes.

Items may be characterized by their set of features and their value to the user and classified by their complexity or scope they require to be evaluated within.
The term ``active users'' is often used to describe the users recommendations are made to.
Interactions of the user with the system are often referred to as transactions. \cite{Ricci:2011jq}

In a mathematical expression, recommender systems aim to provide for a given user $c \in C$ the recommended item
$$s'_c=\underset {s\in S}{\arg \max}\;{u(c,s)}$$
where $u: C \times S \rightarrow R $ is a utility function measuring the usefulness of item $s \in S$ to user $c \in C$, where $R$ is a totally ordered set (like real numbers $\mathbb{R}$ in a certain range, e.g.).
A recommendation task for an evaluation budget $n$ is the problem of computing the best recommendation $s'_c$ for user $c$ while using just $n$ evaluations of the utility function $u$. As usually $n \ll |S|$, the recommendation task has to approximate the result $s'_c$. Different methods have been developed to extrapolate $u$ or build a meta-model to make useful suggestions $s'$ \cite{Adomavicius:2005ec}.

\paragraph{Content-Based Approach.}
This approach recommends items based on their similarity to items previously rated positively by the user.
For computing the similarity of items, generally, a comparison of their features is used \cite{Ricci:2011jq}.


Genetic algorithms have been used to optimize suggestions according to user profiles \cite{Sheth:bm}.
Similar to the approach we propose, the incorporation of genetic algorithms allows for dynamic adaptation to changing user interests by optimizing filtering agents.
Another method is using the set of rated items to train a Bayesian classifier to predict the usefulness of yet unrated items \cite{Pazzani:1997ut}.

The most common method used for keyword retrieval, called term-frequency/inverse-document-frequency (TF/IDF), at its core weighs occurring words by their frequency \cite{Adomavicius:2005ec}.
As an optimization, an approach based on minimum description length (MDL) has been suggested:
MDL provides a framework to minimize the model's complexity by reducing the number of extracted keywords while retaining the items' discriminability~\cite{Lang:1995fh}.

\paragraph{Collaborative Approach.}
Considered to be the most popular method, collaborative filtering recommends items that have already been rated positively by users with a related taste \cite{Ricci:2011jq}.
In contrast to the content-based approach being item-centered, this strategy could be described as user-centered, clustering users with a similar rating history into peers or virtual communities \cite{Hill:1995df}.
According to \cite{Breese:1998wl}, collaborative filtering algorithms can be classified into two types:

\begin{itemize}
\item \emph{Memory-based} algorithms generally predict the rating of a yet unknown item by calculating a weighted sum of this item's rating by other users, where the weight reflects the similarity of those users to the active user.
Therefore different distance measures have been applied, for example, the Pearson correlation coefficient as used in the Group Lens project \cite{Resnick:1994dq}.
In an alternative approach, users are treated as vectors; their similarity can then be measured by calculating the cosine of the angles between them \cite{Breese:1998wl}.
\item \emph{Model-based} algorithms generally employ probability expressions, measuring the probability of a specific rating by the user.
Therefore Bayesian classifiers can be used for creating clusters, or Bayesian Networks can be employed \cite{park2006context}.
\end{itemize}

\paragraph{Further Approaches.}
Besides these two approaches, knowledge-based, community-based and demographic methods have been suggested \cite{Ricci:2011jq}.
Also, hybrid recommender systems as presented in \cite{Balabanovic:1997dm} have been shown to overcome some of the approaches' weaknesses by combining them.


\subsection{Surrogate-Assisted Genetic Algorithms}

This kind of genetic algorithms is applied to problems where an explicit fitness function does not exist, for example in interactive scenarios, but also to areas where the computational costs of the fitness function would be too expensive. When replacing the real fitness function by the use of approximation, however, the accuracy is generally correlated negatively with the computational cost. \cite{eiben2003introduction}

\paragraph{Evolution Control}
In contrast to the early stage of research, where solely the surrogate model was used for evaluation, surrogates are commonly used in combination with the real fitness function as far as possible, in order to prevent the  convergence towards wrong optima introduced by the surrogate.
Methods for this distribution are often referred to as model management or evolution control. Those can be divided into three categories: \cite{Jin:2011fk}
\begin{itemize}
  \item In \emph{individual}-based control, each generation some individuals are evaluated using the real fitness function while the rest is evaluated using the surrogate. Re-evaluating the best-approximated individuals using the real fitness function has been shown to further reduce computational costs \cite{Jin:2011fk}. Another approach applied in \cite{Jin:2004hq} utilizes clustering to evaluate only the most representative individuals.
Re-evaluating the most uncertain predictions has also been proven to be useful, increasing the meta-model's prediction precision by exploring still less evaluated areas.
As measures for the prediction accuracy, simple distance-based techniques, as well applications of Gaussian processes have been proposed \cite{Jin:2011fk}.
  \item \emph{Generation}-based control evaluates some generations entirely using the real fitness function, while the other generations are approximated.
  \item \emph{Population}-based control employs coevolution with multiple populations using different meta-models, while the migration between populations is allowed to individuals. A homogeneous incorporation using neural network ensembles, benefiting from diverse predictions by those has been proposed in \cite{Jin:2004hq}.
Also, heterogeneous methods, utilizing a population-based model management to employ surrogates of different fidelities have been investigated \cite{Sefrioui:2000fe}.
\end{itemize}

\begin{figure*}[t]
  \centering
  \includegraphics[width=0.88\textwidth]{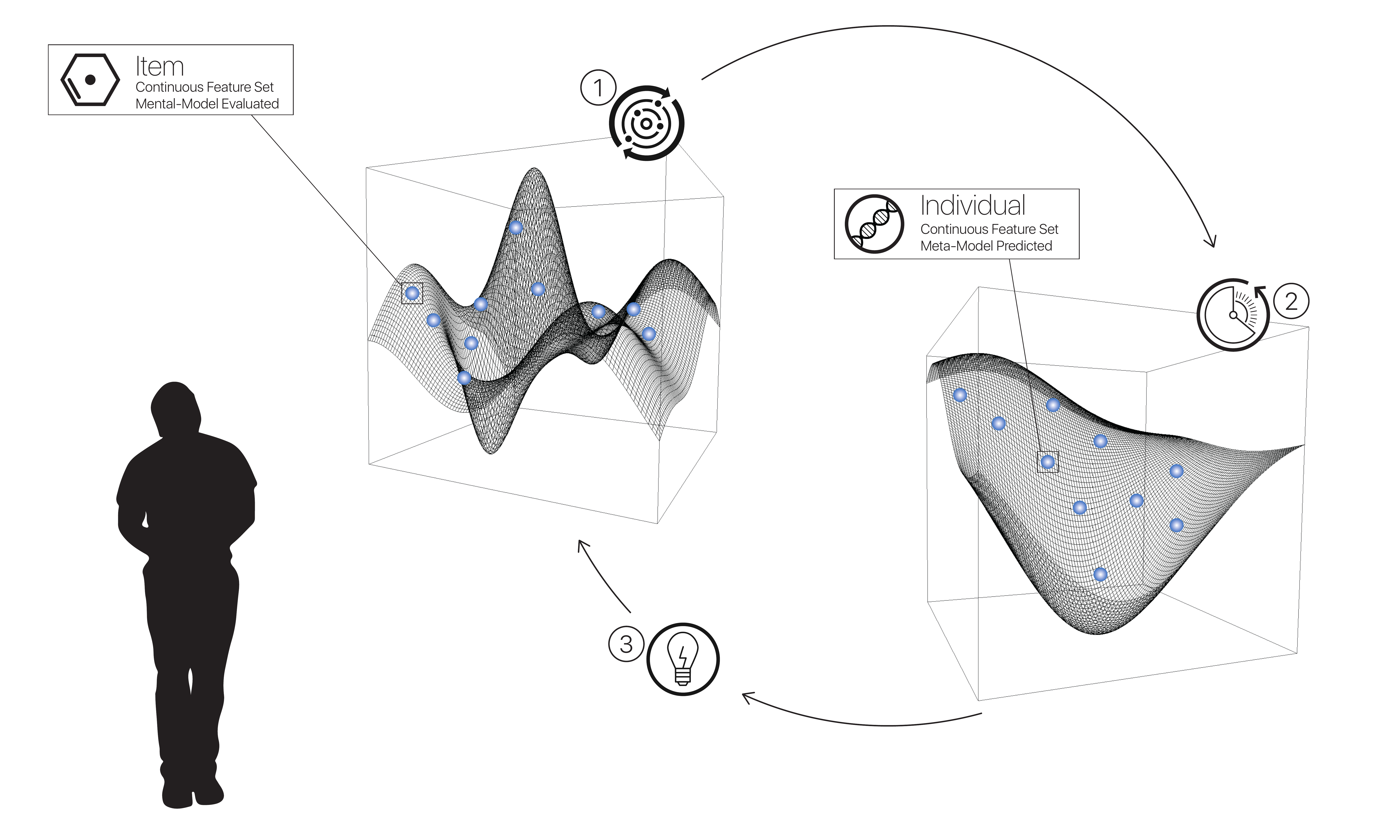}
  \caption[Surrogate-assisted genetic recommender system]{The surrogate-assisted genetic recommender system. A pool of user-evaluated items is used to train a meta-model (1), which in turn is used as a surrogate fitness function in a genetic algorithm optimizing recommendations (2). Once suggest-able and potentially valuable individuals are optimized, they are suggested to the user, evaluated against his or her mental-model and the meta-model is updated to optimize new suggestions (3). This cycle is carried on until the recommendations converge, i.e., no more new or valuable items are found.}
  \label{fig:GRS}
\end{figure*}

It has also been shown that prevention of convergence towards false optima that could be introduced by the surrogate's prediction errors needs to be considered \cite{YaochuJin:2002gn}.

While mainly used for fitness approximation, surrogates have also been applied to the population initialization, mutation, and crossover guiding those otherwise probabilistic mechanisms \cite{Jin:2005bp}.

Regarding data sampling both off-line techniques, i.e., training the meta-model before the optimization, and on-line techniques, i.e., continually updating the surrogate during the optimization process, have been applied \cite{Jin:2005bp}.

\paragraph{Meta-models}
Methods for constructing meta-models can generally be divided into three categories by their level of approximation.
Problem approximation, i.e., trying to replace the original problem with a similar problem which is easier to solve, functional approximation, i.e., trying to simplify or approximate the original fitness function, and evolutionary approximation, specific to evolutionary algorithms containing methods like fitness inheritance or the clustering strategy mentioned above \cite{Jin:2005bp}. As concrete approximation models, different polynomial models, Kriging models, support vector machines and neural networks have been suggested \cite{Jin:2005bp}.

The approach presented in \cite{Kamalian:2007cv} follows our motivation and employs interactive evolutionary computation, combined with a meta-model to embed the human expert domain knowledge, solving computationally expensive modified nodal analysis.

To compensate for the lack of training data available in the use of interactive genetic algorithms in conjunction with surrogate-assisted fitness approximation, the approach presented in \cite{XiaoyanSun:2013gz} employs co-training of radial basis function networks in a homogeneous multi surrogate fashion.
Furthermore, they attempt to handle uncertainties of human interval-based fitness evaluations using a best-strategy individual-based model management.

Besides the application areas mentioned above, surrogate-assisted genetic algorithms have also been found to be helpful for dynamic optimization, constrained optimization and applied to higher optimization robustness \cite{Jin:2011fk, Jin:2005cm}.

\section{Approach}
\label{sec:Approach}

The approach we present can be classified as a content-based recommender system, using two different model-based techniques for filtering, where the results are optimized using a aurrogate-assisted genetic algorithm (SAGA).
The system is built cyclically in order to improve itself among the suggestions and adapt to changes in the user's taste.
See \autoref{fig:GRS} for visual reference depicting the system architecture as motivated by the use case of human-centered recommender systems.

For a given user $c \in C$, we assume that user $c$ can evaluate any item $s \in S$ according to his liking. This evaluation is described by the utility function $u_c : S \rightarrow \mathbb{R}$, assuming that the user's liking can be encoded in $\mathbb{R}$. We also call $u$ the \emph{mental-model} of the user as it describes the user's true intent.

Note that for the purpose of this paper, we are more interested in testing the capabilities of the algorithm than testing user interaction. Thus, for the mental-model $u_c$ we employ well-known and well-defined functions commonly used for benchmarking evolutionary algorithms, i.e., the Bohachevsky, Ackley and Schwefel functions. Accordingly, our item space is $S = \mathbb{R}^2$ to fit these functions \cite{benchmarks}.

Given a set of already evaluated items $S' \subset S$, the goal of our recommender system is to provide us with an item recommendation $$s^* \approx \underset {s \; \in \; S \setminus S'}{\arg \max}\;{u_c(s)}$$ that is the best item not yet discovered by the user. However, we aim to approximate that item without actually calling $u$ and thus allow for some error regarding optimality. Instead we return $$\hat{s} \approx \underset {s \; \in \; S \setminus S'}{\arg \max}\;{\hat{u}_{c}(s, S')}$$ for a surrogate utility function $\hat{u}_c : S \times 2^S \rightarrow \mathbb{R}$, which we call the \emph{meta-model}. The meta-model allows us to describe the suspected quality of items without actually evaluating them with respect to $u$. Note that in order to minimize errors, the meta-model may take into account all the already evaluated items $S'$. The defining feature of this recommendation-based instance of a SAGA is that the best recommendations according to the meta-model are then subsequently evaluated using the mental-model, i.e., $S' \leftarrow S' \cup \{\hat{s}\}$, and thus removed from the search space $S \setminus S'$.

As meta-models, a polynomial regression model fitted to the training data using the method of least squares, and an interpolation model utilizing radial basis function networks are tested and compared in this paper:

\begin{itemize}

\item  For the surrogate model based on \emph{polynomial regression} we employ the second order polynomial $\hat { y } =\beta _{ 0 }+\sum _{ 1 \leq i \leq n }^{  }{ \beta _{ i }x_{ i } } +\sum _{ 1 \leq j \leq n }{ \beta _{ n+j }x_{ j }^2  } $ as suggested in \cite{Jin:2005bp}. The model is fitted to the training data, i.e., the $100$ currently evaluated items using the \emph{least squares method}, which is why we shortly refer to this surrogate as ``LSM''. Thus, given the $2d+1 \times n$ input matrix $X$ and the $n \times 1$ response matrix $Y$, derived from $S'$, the fitness $u_c(s)$ for any unknown item with the $d$-dimensional value vector $s$ can then be estimated by $\hat{u}_c(s, S') = s \; \hat \Theta$  with $ \hat \Theta =  \left( X^TX\right)^{-1}X^T Y $.

\item To build an \emph{interpolation}-based model we use \emph{radial basis function networks} and refer to this model as ``RBF''. As the activation function we utilize the minimization adopted Gaussian function $\phi(x) = 1 - e^{-\left( \frac{x^2}{2\sigma^2} \right)}$.

Given a set of $n$ input vectors and the target vector $T=\left( t_1, ..., t_n \right)$, any unknown item's fitness can be approximated by $\hat{u}_c(s, S') = \sum_{n=1}^{N}{w_n \phi \left( \left\| s-s_n \right\| \right)}$ with $W=\Phi^{-1} T$, where $$\Phi =
	\begin{bmatrix}
		\phi\left( \left\| s_1-s_1 \right\| \right) & \cdots & \phi\left( \left\| s_1-s_n \right\| \right) \\
		\vdots  & \ddots & \vdots \\
		\phi\left( \left\| s_n-s_1 \right\| \right) & \cdots & \phi\left( \left\| s_n-s_n \right\| \right)
	\end{bmatrix},$$ using the euclidean norm $\left\| p - q \right\|_2$ as a distance function.

For finding suitable values for $\sigma$, we use a diversity-based method as suggested in \cite{Bishop:1995wd}, setting the width of all activation functions to the average distance between the currently evaluated items' values.
This method results in a wider prediction landscape when the item pool is diverse and a narrower evaluation landscape once the evaluated items start converging towards an optimum.

\end{itemize}


Individuals to be recommended are selected using the best-strategy, i.e., individuals with the highest predicted fitness value according to $\hat{u}$ are recommended, thus added to the item pool $S'$, and evaluated according to $u$. The amount of suggested individuals per cycle is a free parameter of this approach (cf. Section~4).

\section{Evaluation}
\label{sec:Evaluation}
\subsection{Benchmark Objectives}
\begin{figure*}[hbt]
\captionsetup[subfloat]{aboveskip=10pt}
  \centering
  \subfloat[Bohachevsky objective]{
   \label{fig:Benchmarks:Bohachevsky}
   \includegraphics[width=0.33\linewidth]{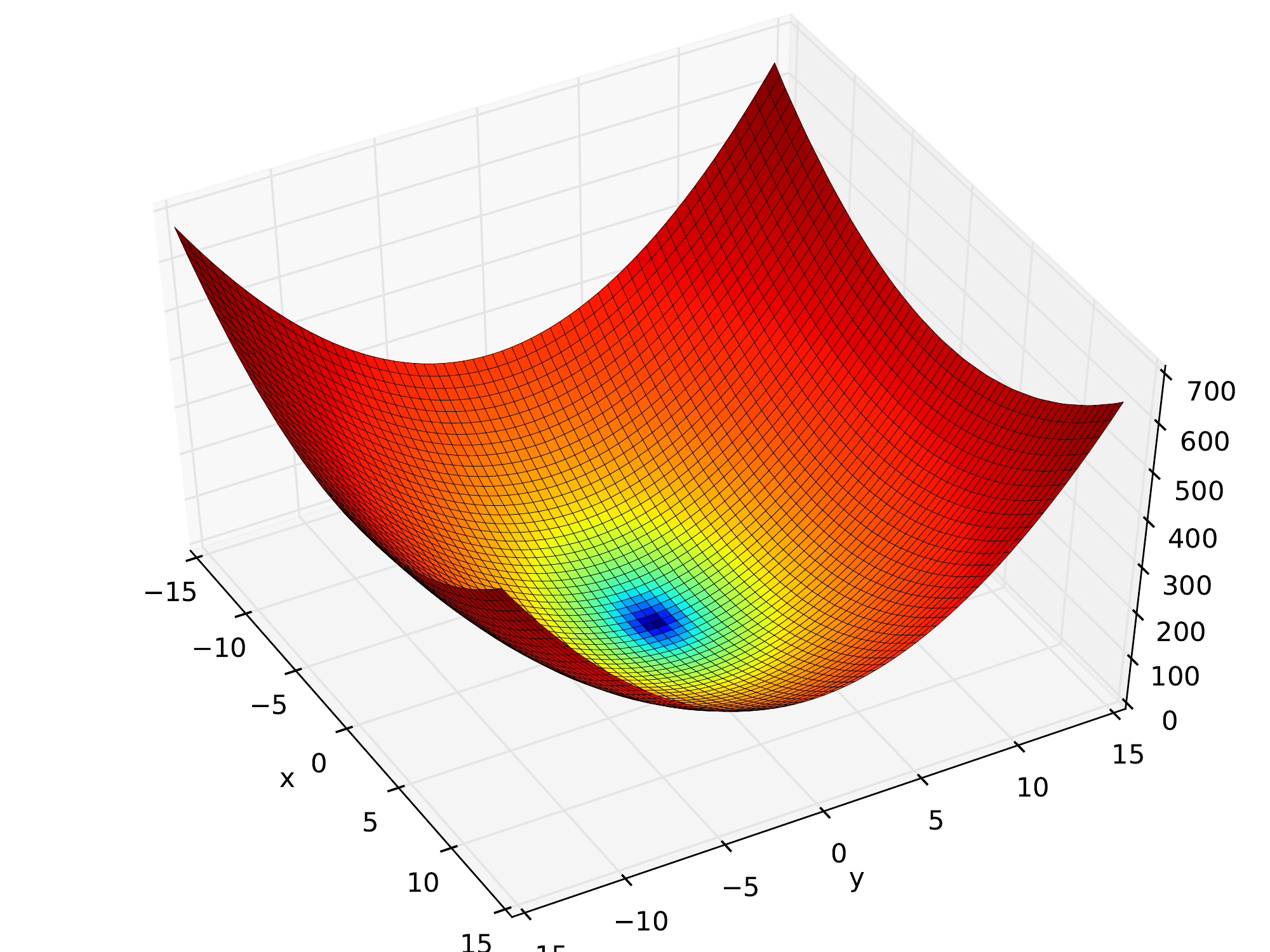}}
  \subfloat[Ackley objective]{
   \label{fig:Benchmarks:Ackley}
\includegraphics[width=0.33\linewidth]{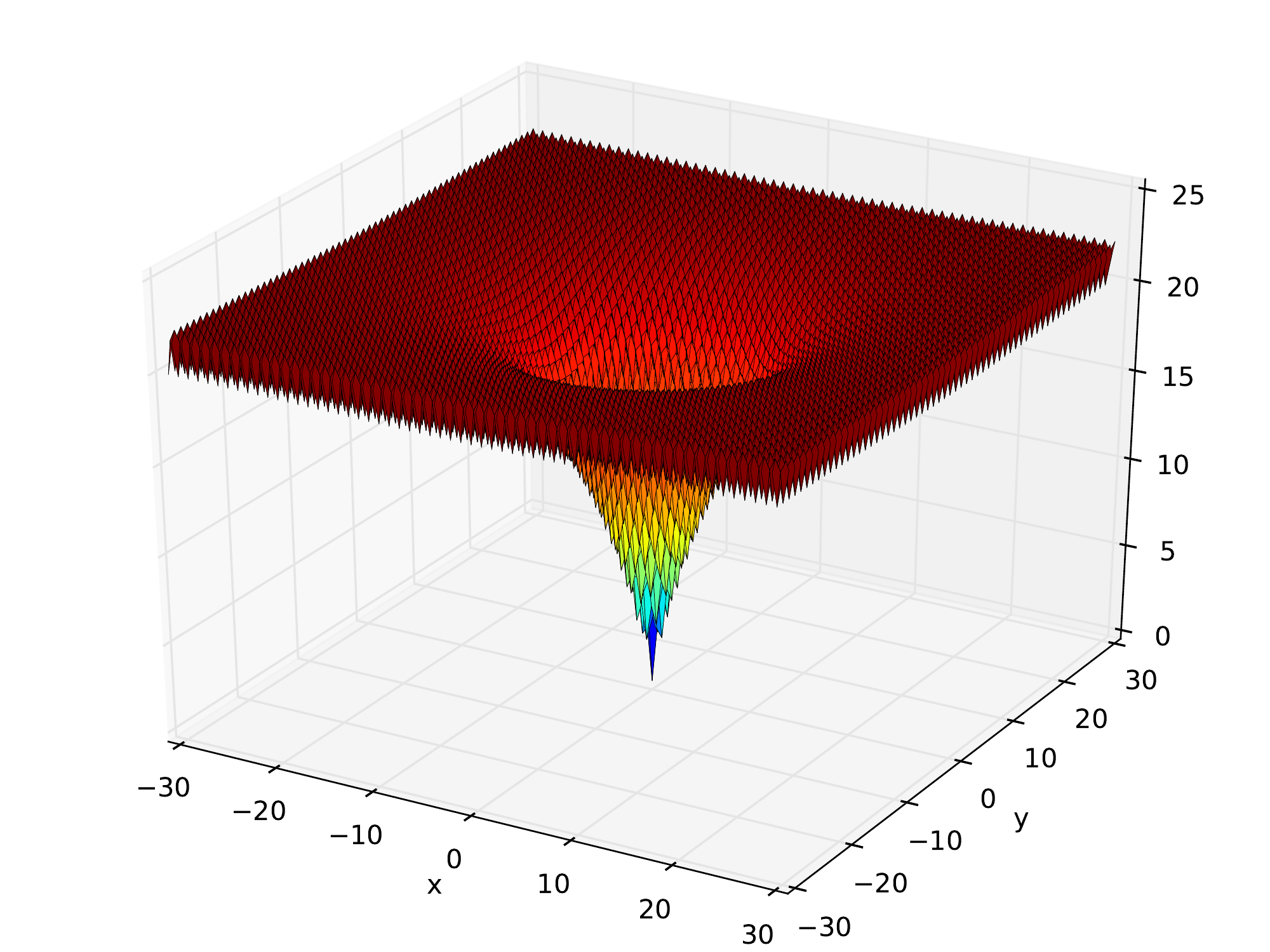}}
  \subfloat[Schwefel objective]{
   \label{fig:Benchmarks:Schwefel}\includegraphics[width=0.33\linewidth]{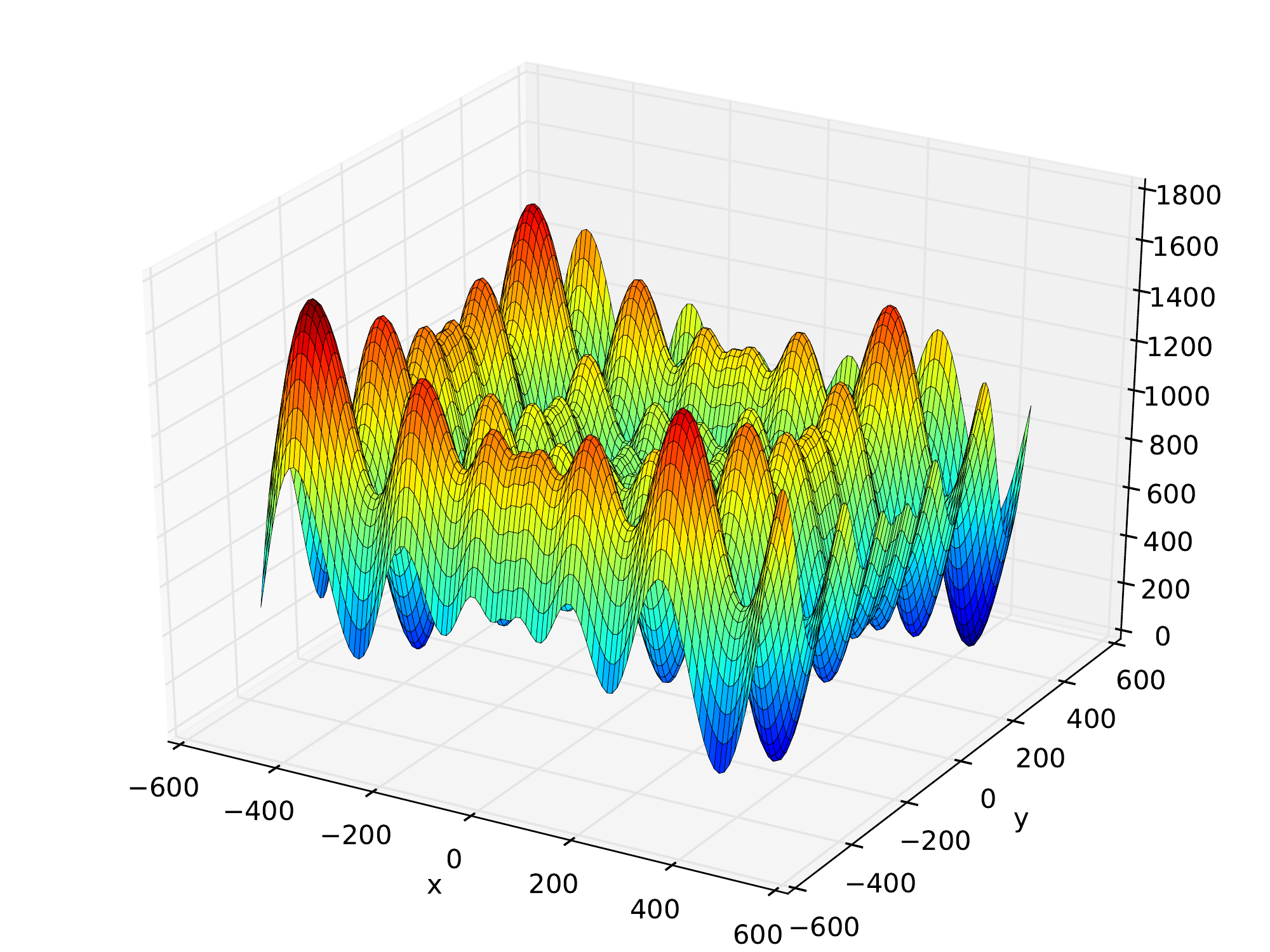}}
  \caption[Fitness Landscapes of Bohachevsky, Ackley, and Schwefel benchmark]{Function plots for the (a) Bohachevsky, (b) Ackley, and (c) Schwefel benchmark functions for two-dimensional input. Images taken from \cite{benchmarks}.}
  \label{fig:Benchmarks}
\end{figure*}

In order to run a multitude of tests on the performance of the employed approach and the employed models, we opt for standard evolutionary benchmark functions instead of real human interaction. We used the implementation for Bohachevsky, Ackley, and Schwefel functions provided by \cite{benchmarks}. See \autoref{fig:Benchmarks} for a small illustration. All of these are constrained to a specific subset of $\mathbb{R}^2$ and are to be minimized with a best fitness value of $0$.

\subsection{Parameter Optimization}
To test and compare different settings, data regarding the best item's fitness and accepted suggestion (in every suggestion cycle) as well as the fitness of the best real-evaluated item and the cycle of convergence (after the fixed number of recommendation cycles) are saved for each test run.
The cycle of convergence is represented by the last recommendation cycle in which at least one suggestion was accepted.
The number of accepted suggestions is obtained by counting the suggestions that are evaluated better than the worst evaluated item at that time. For enhanced comparability, the number of accepted suggestions is then normed with the number of suggestions, so that this variable displays the success in the range $[0;1]$.
The following section will provide tests and evaluations regarding the population-handling technique, the rate of evaluation, the amount of suggestions, and the optimal number of recommendation cycles.
Every test setup is performed in $10$ repetitions and the results are averaged in order to get a more representative result, less influenced by possible outliers, also illustrating the robustness.
To keep the computational efforts reasonably low, all the tests are made based on two-dimensional versions of the objectives introduced above.

\subsubsection{Rate of Evaluation and Population-Handling Technique}
\begin{figure*}[t]
  \centering
  \includegraphics[width=\textwidth]{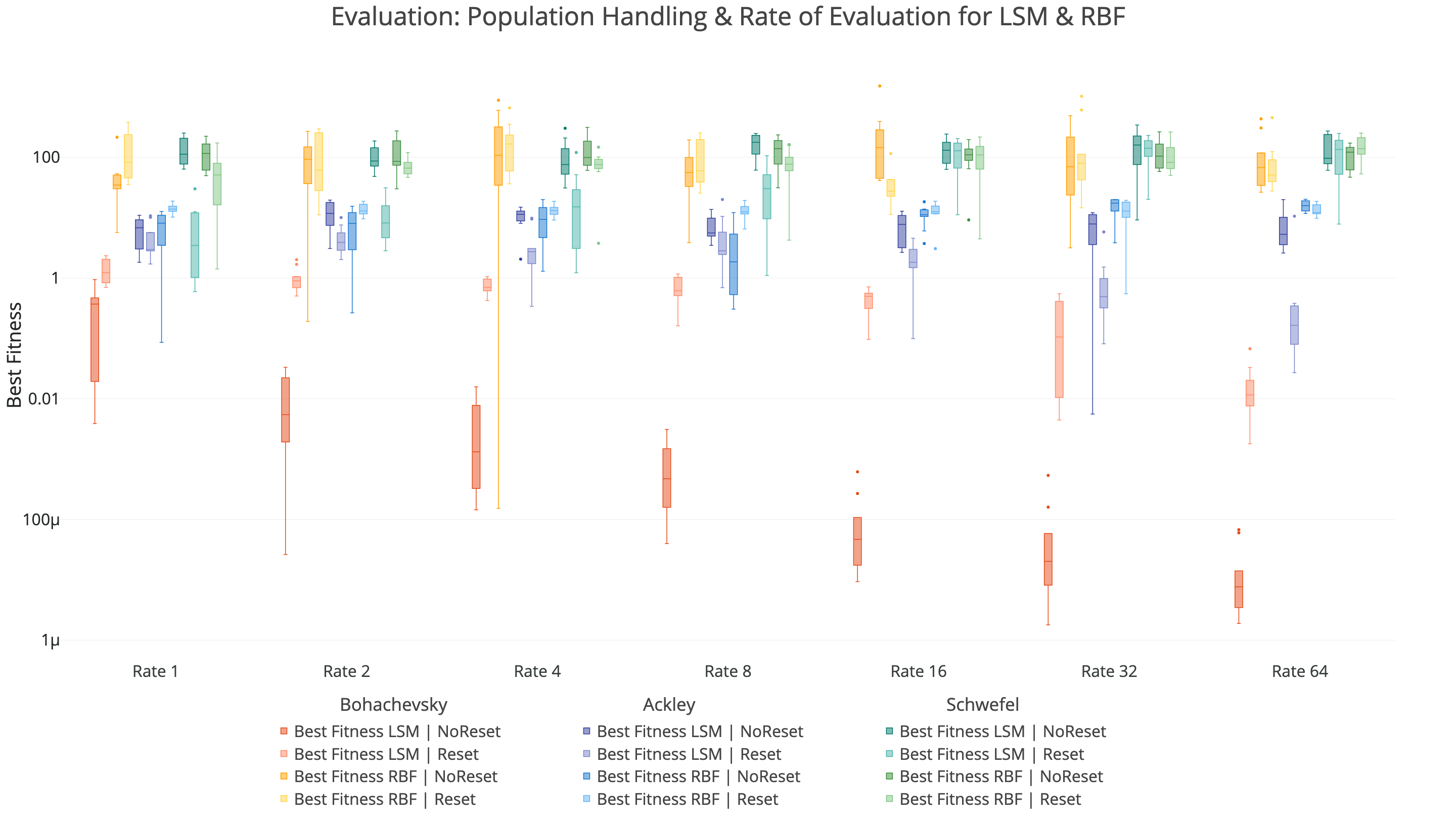}\\
  \caption[Benchmark Rate of Evaluation and Population-Handling Technique]{Benchmarking different Evaluation Rates (x-axis) on the Bohachevsky (red/orange), Ackley (blue) and Schwefel (green) objective. The LSM- and RBF-Model  are compared with a Reset (lighter color) and NoReset (darker color) population-handling technique. Test results for ten iterations of each test setup are displayed by box-plots for the Best Fitness on a logarithmic scale.}\label{fig:rates}
\end{figure*}

Test results on different evaluation rates, i.e., number of optimization cycles before suggesting, ranging from $1$ to $64$, as well as the impact of resetting the pool of individuals or maintaining that pool throughout the recommendations are visualized in \autoref{fig:rates}.
%
%

Overall, the test results imply that applying a no-reset population handling strategy yields a better outcome if the model is not at risk to converge towards a local optimum of the benchmark (see results for the Bohachevsky benchmark).
A counterexample for this can be observed at the results for the LSM model on the Ackley benchmark.
Also, choosing lower rates, i.e., shorter optimization of suggestions, further reduces the risk of introducing false optima to the surrogate by decelerating the model's convergence.
Otherwise, higher rates are able to benefit the system's performance, as seen at the rest results for the LSM model on the Bohachevsky benchmark.

\subsubsection{Amount of Suggested Individuals per Cycle}
\begin{figure*}[t]
  \centering
  \includegraphics[width=\textwidth]{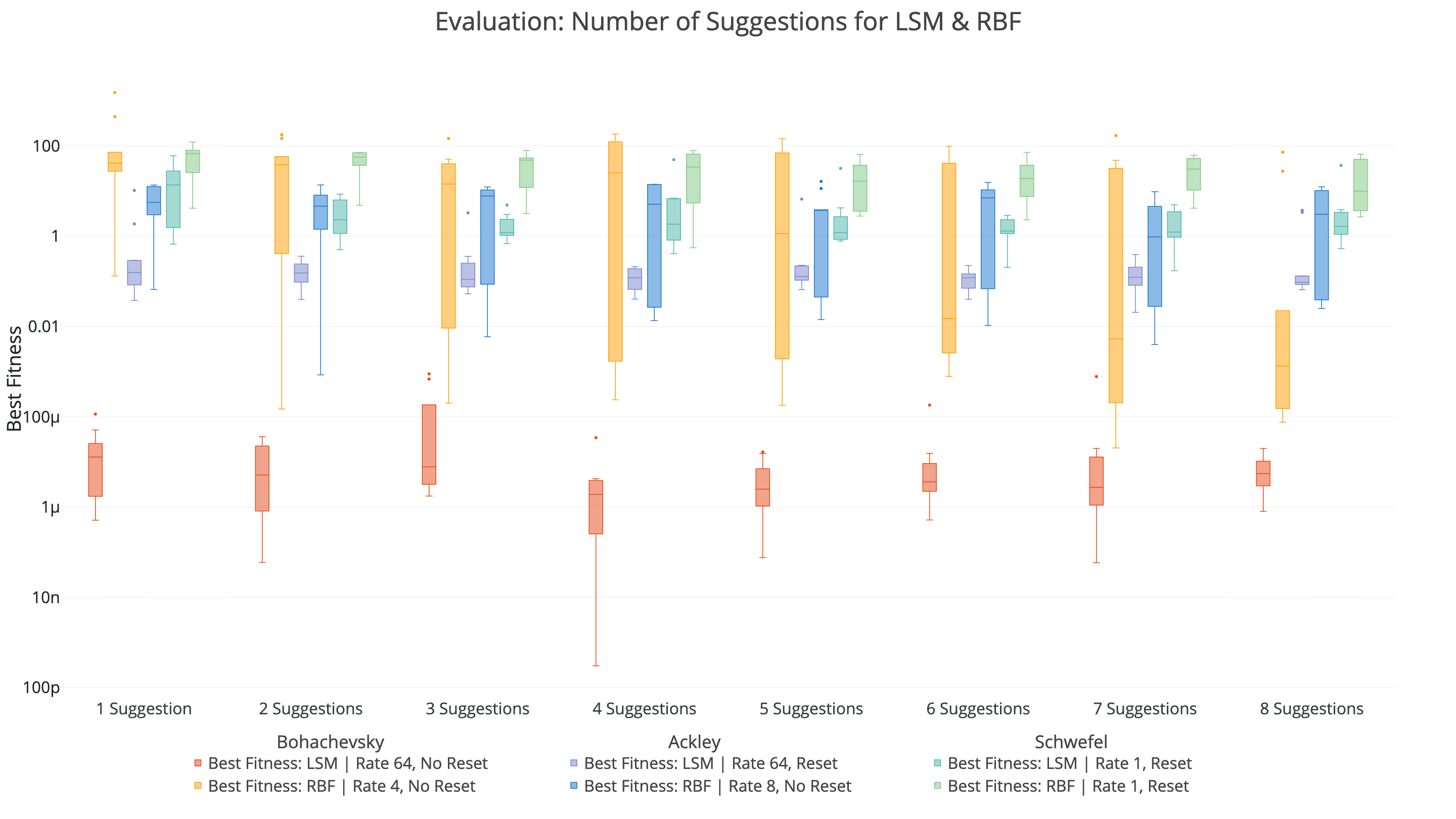}\\
  \caption{Testing the impact of the number of suggestions (x-axis) per recommendation cycle on the Bohachevsky (red/orange), Ackley (blue) and Schwefel (green) objective. Comparing the LSM and RBF models by their Best Fitness, displayed by box-plots on a logarithmic scale. Concrete parameter settings for the Evaluation Rate and the population-handling technique are shown in the legend below.}
  \label{fig:suggestions}
\end{figure*}

Regarding the amount of recommendations per cycle, numbers from 1 through 8 have been tested; test results can be seen in \autoref{fig:suggestions}.
Overall, a higher number of suggestions benefits the systems performance, especially in combination with shorter optimization of those, as this leads to a higher diversity of recommendation, which could counteract the risk of the model's convergence towards a local or false optimum.
Also, the comparably local perspective of the RBF model, causing an overall worse performance than the LSM model, seems to be neutralized by this effect, as seen in the results for the RBF model on the Bohachevsky benchmark.
The results for the LSM model on both the Bohachevsky and Ackley benchmarks show less correlation, as the results are already quite good and further improvement would be hard to achieve, especially with this more globally oriented surrogate model.
%
%

\subsubsection{System Convergence and Number of Recommendation Cycles}
\begin{figure*}[t]
  \centering
  \includegraphics[width=0.96\textwidth]{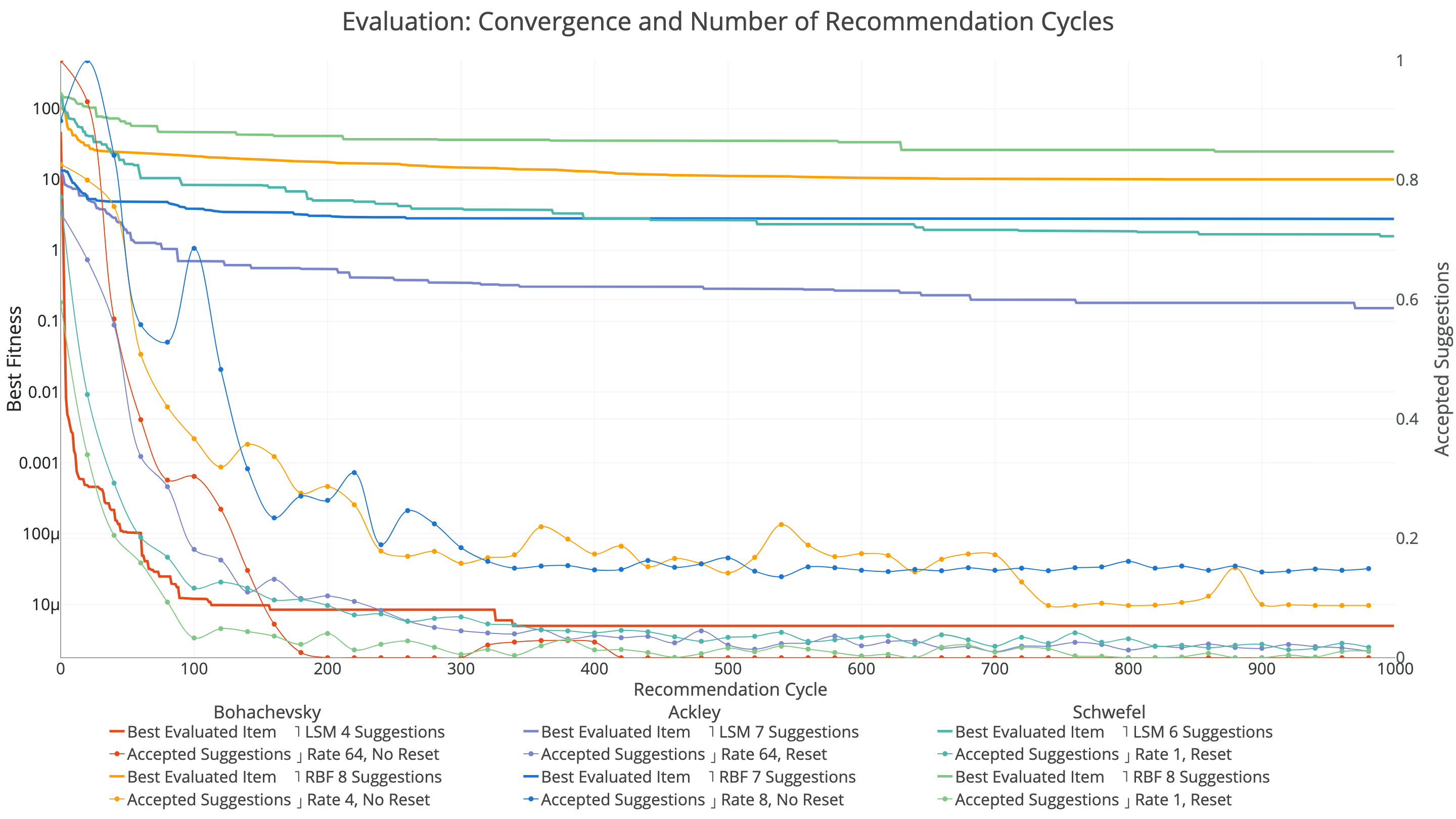}\\
  \caption[Evaluating the amount of recommendation cycles needed for both surrogates on the Bohachekvsky objective ]{Evaluating the optimal number of recommendation cycles (x-axis) needed for both LSM and RBF models on the Bohachevsky (red/orange), Ackley (blue) and Schwefel (green) objective. The optimization progress represented by the Best Fitness (left y-axis) is visualized by a line graph on a logarithmic scale, the share in accepted suggestions (right y-axis) represented by dotted line graphs. Concrete parameter settings for the Evaluation Rate, population-handling technique, and number of suggested items per recommendation cycle are displayed in the legend below.}
  \label{fig:convergence}
\end{figure*}

As mentioned above, the tested system is intended to work in an interactive scenario; thus we strive for an overall low amount of real fitness evaluations.
In order to further reduce those, additional tests have been carried out to determine the minimal amount of recommendation cycles required.
The convergence of the system, i.e., if it is still able to make valuable recommendations is measured by the number of accepted suggestions while the performance of the surrogate and its convergence can be derived from the fitness of the currently best real-evaluated item.
The test results visualized in \autoref{fig:convergence} show that all of the systems mostly converge within about $100$ cycles, which, depending on the specific settings requires about $1000$ real fitness evaluations at maximum.
Reaching further improvement with a higher number of recommendation cycles could not be justified by the amount of additional real evaluations that would be needed.

%
%
%

%

\subsection{Comparison of System Performance}

In order to put the Surrogate-Assisted Genetic Recommender System's (``SAGRS'') performance into a context, we compare it to the conventional Genetic Algorithm (``GA'').
For all evolutionary processes (both within the SAGRS and for the GA itself), a selection factor of $0.9$, a mutation probability of $0.1$, and a recombination probability of $0.05$ are used.
Furthermore an equal amount of true fitness evaluations $n_\textit{eval}$ is retained, defining the population size $n_\textit{pop}$ and the number of generations $n_\textit{gen}$ in even distribution, such that $n_\textit{pop} = n_\textit{gen} = \left \lfloor{\sqrt {n_\textit{eval}}}\right \rfloor$.
If compared to LSM and RBF models with different amounts of true evaluations, the higher one is used to compute the settings.

To further validate the necessity and the performance of the meta-models used and to get an impression of the impact the genetic optimization has, comparisons against a random search strategy (Random Recommender, ``RR''), integrated into the system the same way the genetic optimization is, are drawn.
Random search is implemented as a genetic optimization with an evaluation rate of 0, causing the optimization to be skipped and the system to recommend the initially best-estimated individuals.
As parameters for this random-search-adopted recommender, an evaluation rate of 0 and a reset population-handling technique as well as the optimal settings for the number of suggestions and the amount of recommendation cycles (as evaluated in the previous sub-section) are used.
From a broader perspective, this system could also be seen as a conventional content-based recommender-system, making suggestions solely based on the estimation of the unknown item's rating.
Concrete settings for all of those variable parameters of each system compared are annotated in the legends of the plots.

\begin{figure}[!h]
  \includegraphics[width=0.5\textwidth]{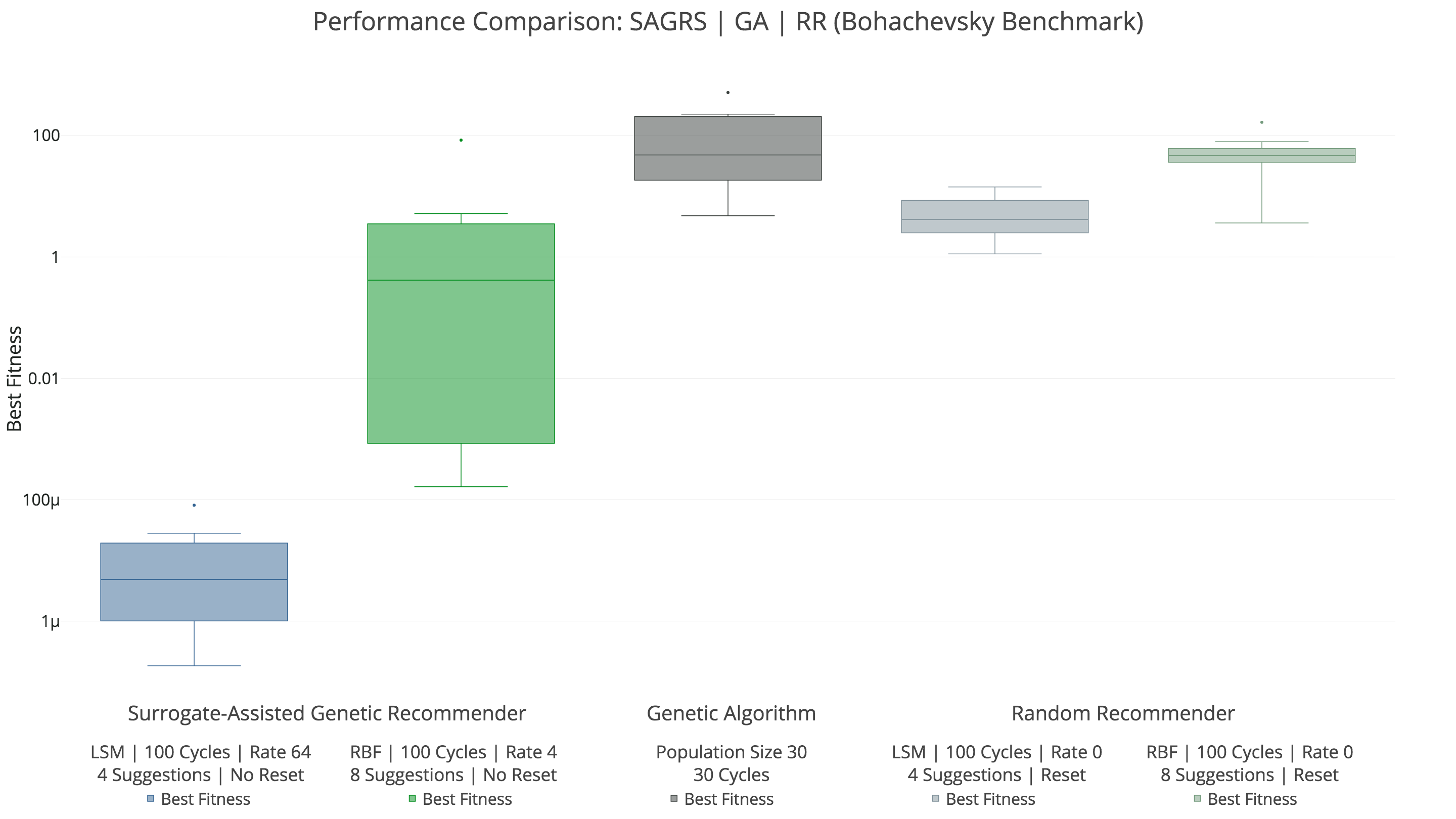}\\
  \caption[Performance comparison results on the Bohachevsky objective]{Performance comparison of the SAGRS using an LSM model (blue) and an RBF model (green), a conventional genetic algorithm (dark grey), and a random recommender based on the LSM (blue-grey) and RBF (green-grey) model on the Bohachevsky objective. Their outcome regarding the best fitness is visualized by box-plots on a logarithmic scale. Concrete parameter settings are shown in the legend.\vspace{2em}}
  \label{fig:Cmp_B}
\end{figure}

\paragraph{Bohachevsky.}
The comparison results seen in \autoref{fig:Cmp_B} show that the SAGRS outperforms the Genetic Algorithm as well as the Random Recommender with both approximation models easily.
The reason for this is most likely that the Genetic Algorithm cannot handle such a low amount of evaluations, i.e., it is not able to offer convergence towards better fitness areas with such small amounts of individuals and optimization cycles.

The Random Recommender, also performing better than the Genetic Algorithm, shows that the incorporation of meta-models has a distinctly positive impact on the system's performance.
Comparing the Surrogate-Assisted Genetic Recommender to the Random Recommender furthermore provides evidence that the optimization of suggestions has an affirmative effect.

The LSM model apparently has the best outcome, but also the RBF model performs surprisingly well.
With 100 recommendation cycles of four suggestions, optimized comparably long from a population that is retained over the runtime, the LSM model clearly yields profit from its similarity to the objective and its global estimation capabilities.
The RBF model, on the other hand, takes advantage from an increased diversity of suggestions, as a result to their short optimization in conjunction with the high amount of recommendations per cycle, allowing for a better exploration of the objective, thus counteracting the model's local perspective.

\begin{figure}
  \includegraphics[width=0.5\textwidth]{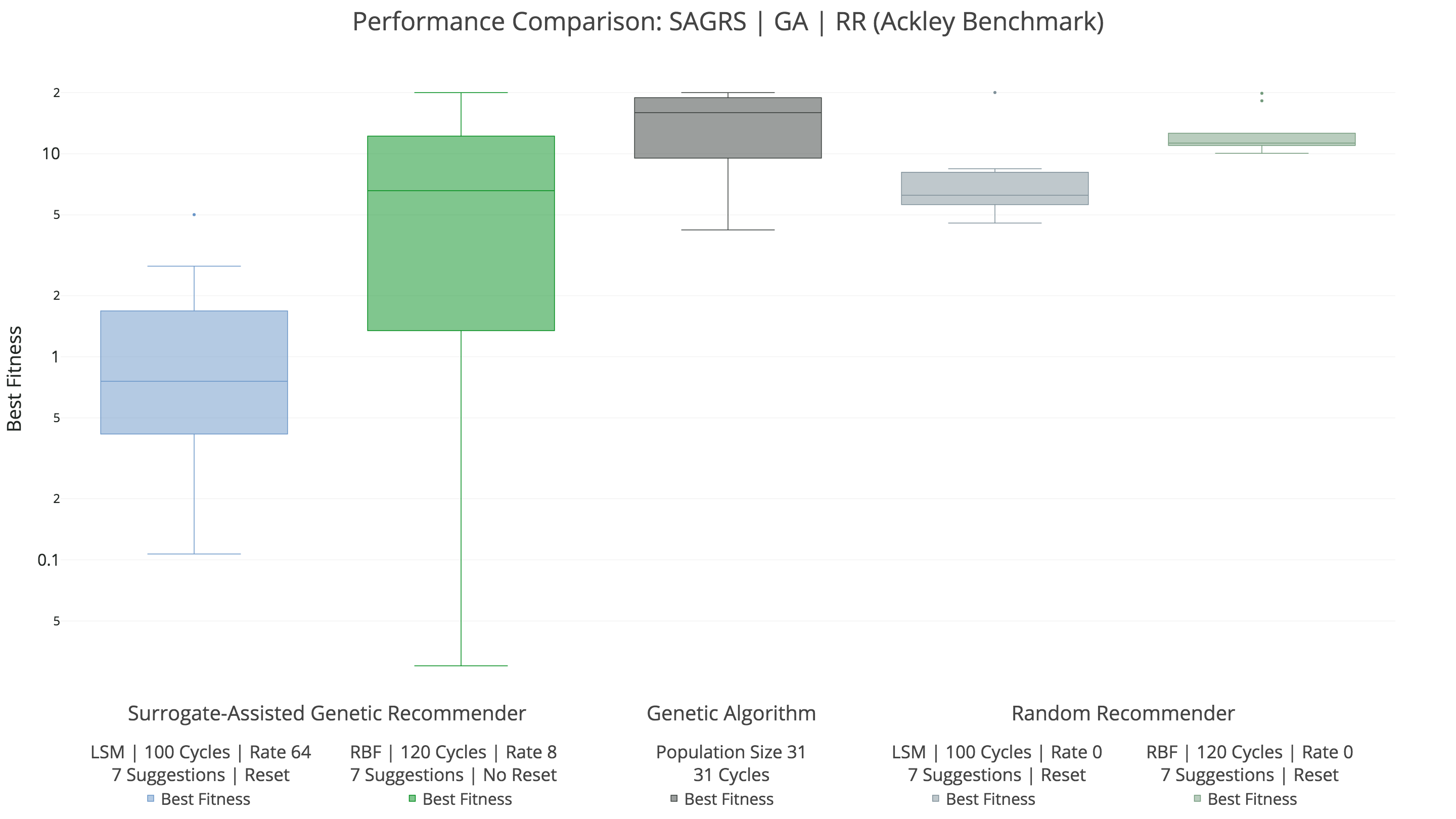}\\
  \caption[Performance comparison results on the Ackley objective]{Performance comparison of the SAGRS using an LSM model (blue) and an RBF model (green), a conventional genetic algorithm (dark grey), and a random recommender based on the LSM (blue-grey) and RBF (green-grey) model on the Ackley objective. Their outcome regarding the best fitness is visualized by box-plots on a logarithmic scale. Concrete parameter settings are shown in the legend.}
  \label{fig:Cmp_A}
\end{figure}

\paragraph{Ackley.}
The comparison results seen in \autoref{fig:Cmp_A} show that, similar to the Bohachevksy objective, the SAGRS is able to outperform the Genetic Algorithm as well as the Random Recommender with both approximation models.
Still, the Genetic Algorithm does not accomplish convergence towards any good solutions with this low amount of fitness evaluations, which seems obvious, given the fact, that Ackley is even harder to be optimized than Bohachevsky, where it already failed.

Even though the results are allocated on a denser expanse, due to the objective's comparably small fitness range, it still can be perceived that the Random Recommender results in a better outcome than the Genetic Algorithm, while being outperformed by the SAGRS.
Therefore, the implications about the usefulness of both the optimization of suggestions and the utilization of approximation models can still be endorsed for this specific benchmark.

While mostly performing worse than the LSM model, the RBF model is able to reach even better results at some times.
With a comparably short optimization and a high number of suggestions, the RBF model can benefit from a higher diversity of recommendations, as seen before.
In contrast, the LSM model is able to prevent being misdirected by the objective's local optima, by not retaining previous populations, therefore requiring longer optimizations, and suggesting a high amount of individuals.

\begin{figure}
  \centering
  \includegraphics[width=0.5\textwidth]{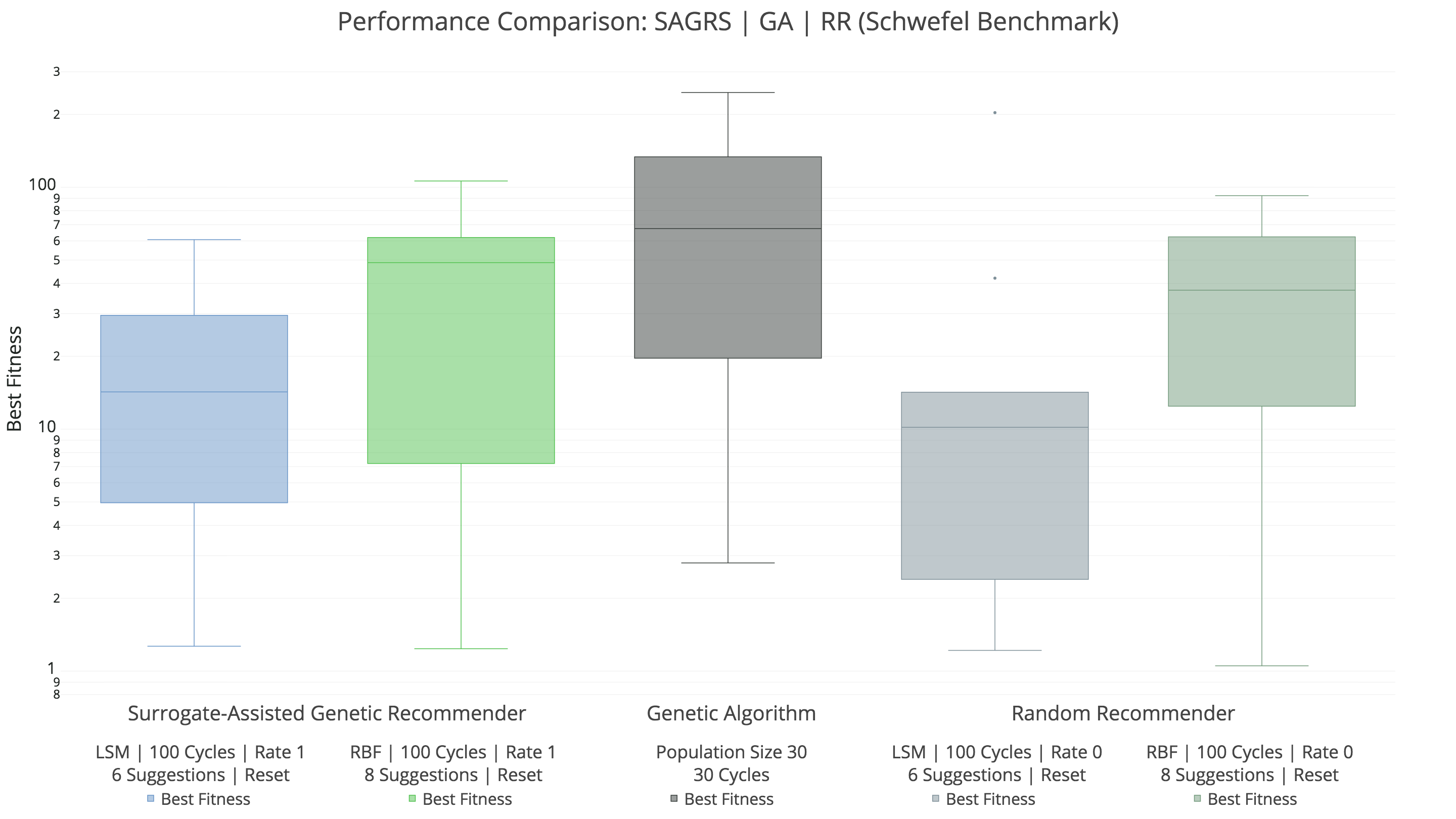}\\
  \caption[Performance comparison results on the Schwefel objective]{Performance comparison of the SAGRS using an LSM model (blue) and an RBF model (green), a conventional genetic algorithm (dark grey), and a random recommender based on the LSM (blue-grey) and RBF (green-grey) model on the Schwefel objective. Their outcome regarding the best fitness is visualized by box-plots on a logarithmic scale. Concrete parameter settings are shown in the legend.}
  \label{fig:Cmp_S}
\end{figure}

\paragraph{Schwefel.}
The comparison results seen in \autoref{fig:Cmp_S} show that the SAGRS is able to slightly outperform the Genetic Algorithm and offers similarly good results compared to the Random Recommender.
After performing badly for the previous two benchmarks, which could be classified as easier than Schwefel, the Genetic Algorithm shows good results despite the low number of evaluations.

Here, the Random Recommender shows a comparably better performance than the SAGRS.
Considering their low evaluation rate of $1$, hardly differing in effect to a rate of 0, and that they the same population-handling technique, such outcome could have been presumed with regard to the test results for the Schwefel objective in the previous section.
Still, the positive impact of the use of approximation models as well as the better performance of the LSM model due to its global estimation capabilities can be seen.

The short optimization, combined with the high amount of suggestions and the reinitialization of the population, causes the convergence of both meta-models to be decelerated to a high degree, preventing them from converging towards the distinct local optima of the objective.

\section{Conclusion}
\label{sec:Conclusion}
We proposed a framework for building systems that are able to make recommendations based on content that has been evaluated by a user. We assumed that user preference can be modeled by a real-valued function called the mental-model, i.e., the true utility to be measured from the user evaluating a given item.
In order to estimate the true utility of items yet unknown to the user, we used a polynomial regression model (fitted using the method of least squares) and an interpolation model (using radial basis function networks) as surrogate models. These were employed by a genetic algorithm to optimize for the best item that has not yet been evaluated according to the mental-model. By evaluating these suggested items on the mental-model, we improve our surrogate model but also force a dynamic change in the mental-model.

We evaluated and tested the approach by replacing the subjective human evaluation with three different objective benchmark functions.
Evaluating and optimizing the impact of the evaluation rate, population-handling technique, number of suggestions and number of recommendation cycles, we realized that decelerating the meta-model's convergence helps to overcome local optima of the objective, and aids the system to converge towards the global optimum.


\paragraph{Limitations.}
As we replaced the human evaluation with benchmark functions, the influences of human evaluation, even though considered, are not tested nor evaluated. 
Therefore an applicability of this approach as an interactive system cannot be stated.
Furthermore, the use of those benchmarks compensates for the need of an appropriate classification of items, which plays an important role when applying the system to real items.
Also, all tests were only made with two-dimensional values, which would be too few for accurately classifying actual items. It should also be noted that while the benchmark functions are well-established for testing genetic algorithms, it is still to be shown if the approach generalizes to other functions beyond that.
When used in a real-world application, especially with a human interaction, the optimized amount of 1000 evaluations would still be a considerably high amount of evaluations to be demanded from a single user.


\paragraph{Future Work.}

As the scope of the performed tests is limited, further tests, especially evaluating and optimizing the system's real-world applicability should be performed.
To test the system's ability adapting to a changing taste of the user, evaluations could be made using a changing fitness landscape for the mental-model, for example, the Moving Peaks benchmark presented in \cite{benchmarks}.
Also, a higher dimensionality of the items' features should be tested to get a more representative image of the performance.
Another test to be performed is the reaction to noisy fitness functions, as a human evaluation might involve uncertainty.

Since the approximation models showed some weaknesses, a further improvement of those should also be considered.
Even though considered to be a more powerful model, the radial basis function network was mostly outperformed by the polynomial regression model, due to a too local point of view, resulting from the interpolation technique.
To counteract these constraints, training the model with fewer radial basis function nodes by clustering the sample data might be helpful.
An alternative approach for constructing a meta-model could utilize a Gaussian process as suggested in \cite{Jin:2011fk}, which already offers the Gaussian mean and variation as a measure of certainty.
Based on the idea of hybrid recommender systems, a combined incorporation of both surrogates might help to overcome some of their weaknesses, especially due to their different level of approximation.
Having shown that the diversity of suggestions is able to influence the model's convergence to prevent the convergence towards local optima, active control over the exploration and exploitation might be useful.
Therefore, a most-uncertain or novelty-based technique for selecting items to be recommended could be used, which would need the approximation models to be extended by a measure of certainty.

Lastly, the approach should be implemented in a real-world scenario to really test the human interaction instead of making assumptions about it.
Most importantly, different methods for incorporating the human evaluation need to be evaluated to provide a helpful tool, assisting its users to deal with the vast variety of possibilities most efficiently.

\vspace{3em}